\documentclass[journal]{IEEEtran}

\usepackage{graphicx}
\usepackage{float}
\usepackage{lineno}
\usepackage{minted}
\usepackage{amsmath}
\usepackage{amssymb}
\usepackage{amsfonts}
\usepackage{booktabs}
\usepackage{cite}
\usepackage[colorlinks=true, citecolor=blue, linkcolor=blue, urlcolor=blue]{hyperref}
\newcommand{\hide}[1]{}
\begin{document}

\title{Reinforcement Learning Agent\\ for a 2D Shooter Game}

\author{Thomas Ackermann$^1$ , Moritz Spang$^2$ , Hamza A. A. Gardi$^{2,3}$ \\
{\small \{$^1$ Faculty for Mathematics, $^2$ Department of Electrical Engineering and Information Technology, $^3$ IIIT at ETIT\},\\ Karlsruhe Institute of Technology, 76131 Karlsruhe, Germany 
}}

\maketitle

\setlength\linenumbersep{4pt}

\begin{abstract}
Reinforcement learning agents in complex game environments often suffer from sparse rewards, training instability, and poor sample efficiency. This paper presents a hybrid training approach that combines offline imitation learning with online reinforcement learning for a 2D shooter game agent. We implement a multi-head neural network with separate outputs for behavioral cloning and Q-learning, unified by shared feature extraction layers with attention mechanisms.

Initial experiments using pure deep Q-Networks exhibited significant instability, with agents frequently reverting to poor policies despite occasional good performance. To address this, we developed a hybrid methodology that begins with behavioral cloning on demonstration data from rule-based agents, then transitions to reinforcement learning.

Our hybrid approach achieves consistently above $70\%$ win rate against rule-based opponents, substantially outperforming pure reinforcement learning methods which showed high variance and frequent performance degradation. The multi-head architecture enables effective knowledge transfer between learning modes while maintaining training stability. Results demonstrate that combining demonstration-based initialization with reinforcement learning optimization provides a robust solution for developing game AI agents in complex multi-agent environments where pure exploration proves insufficient.
\end{abstract}


\IEEEpeerreviewmaketitle

\section{Introduction}
Reinforcement Learning (RL) has emerged as a foundational paradigm in artificial intelligence for training agents to make sequential decisions through interaction with dynamic environments. By optimizing behavior through reward feedback rather than explicit supervision, RL enables the development of systems capable of autonomously learning complex tasks, a characteristic crucial for applications in robotics, autonomous systems, personalized healthcare, finance, and more \cite{sutton2018reinforcement, li2017deep}. 

One of the most significant advantages of RL is its general applicability to real-world problems where the optimal policy is not easily prescribed. For example, in robotics, RL facilitates the learning of control policies for tasks such as manipulation and locomotion in unstructured environments \cite{kalashnikov2018qt}. In healthcare, RL algorithms have shown promise in optimizing treatment strategies for chronic diseases by adapting to individual patient responses over time \cite{komorowski2018artificial}. \\
Despite these promising directions, the deployment of RL in the real world is often hindered by challenges such as sample inefficiency, lack of interpretability, and the high cost of failure during training. As a result, video games have become an essential tool for advancing RL research. Games offer rich, interactive environments that are safe, scalable, and well-suited for benchmarking algorithmic progress \cite{bellemare2013arcade, kempka2016vizdoom}. Notable breakthroughs have demonstrated the potential of RL in complex game environments. DeepMind's AlphaGo achieved superhuman performance in the ancient board game Go, combining deep neural networks with Monte Carlo Tree Search (MCTS) and RL to defeat world champions \cite{silver2016mastering}. Building on this, AlphaZero further demonstrated that RL alone can be used to master multiple board games without human data \cite{silver2018general}. 

These accomplishments highlight the value of games as controlled experimental platforms where RL agents can learn to operate in high-dimensional, partially observable, and stochastic environments. Insights gained from these domains are now being transferred to robotics, autonomous driving, and other high-stakes fields, where generalization, robustness, and real-time adaptability are essential. \\\\
In this paper, we explore strategies for implementing an agent through RL for a 2D shooter game developed in Python. With the help of state of the art strategies, we aim to find the optimal strategy, so that the win rate of the agent in the game is maximized. A game is won, when the agent, who serves as a player, is able to defeat the enemy on the map by shooting him a total of 3 times while simultaneously being shot less than 3 times. 

The player and the enemy entity have the same set of actions available to them. After training is completed, the RL agent shall serve as an enemy in the shooter game. Thus, playing the game on your own against an advanced enemy artificial intelligence becomes fun and challenging. \\\\
To find the optimal strategy for implementing the RL agent, our approach is as follows:
\begin{enumerate}
    \item Analyze state of the art strategies for implementing RL agent in shooter games in the current literature
    \item Selecting the best suited and efficient strategies for implementation for teaching the RL agent how to play the game
    \item Implementing those strategies in the context of the shooter game
    \item Evaluate the outcome of the strategies regarding maximizing the win percentage
    \item From the evaluation, choose a suitable strategy for the RL agent in the 2D shooter game
\end{enumerate} 
This approach also reflects the structure of this paper. In the end, we will provide a conclusion of our approach and our findings and give an outlook, on what can be improved or rather adapted for future works.

\section{Fundamentals}
\subsection{Reinforcement Learning}

Reinforcement Learning (RL) is a computational framework in which an agent learns to make decisions by interacting with an environment to maximize cumulative rewards. This process is formalized as a Markov Decision Process (MDP) defined by the tuple $(\mathcal{S}, \mathcal{A}, P, R, \gamma)$, where:
\begin{itemize}
    \item $\mathcal{S}$ is the set of states,
    \item $\mathcal{A}$ is the set of actions,
    \item $P(s'|s, a)$ is the transition probability from state $s$ to $s'$ under action $a$,
    \item $R(s, a)$ is the reward function,
    \item $\gamma \in [0,1)$ is the discount factor.
\end{itemize}

The agent’s objective is to find a policy $\pi(a|s)$ that maximizes the expected return, defined as:

\begin{equation}
G_t = \sum_{k=0}^{\infty} \gamma^k R_{t+k+1}
\end{equation}

The value function $V^\pi(s)$ estimates the expected return starting from state $s$ and following policy $\pi$:

\begin{equation}
V^\pi(s) = \mathbb{E}_\pi \left[ \sum_{k=0}^{\infty} \gamma^k R_{t+k+1} \,|\, S_t = s \right]
\end{equation}

Similarly, the action-value function $Q^\pi(s, a)$ estimates the expected return from state $s$ taking action $a$:

\begin{equation}
Q^\pi(s, a) = \mathbb{E}_\pi \left[ \sum_{k=0}^{\infty} \gamma^k R_{t+k+1} \,|\, S_t = s, A_t = a \right]
\end{equation}

In order to calculate the expected future value from subsequent states, the following equation recursively defines the expected value of being in a state $s$, following a specific policy $\pi$.

\begin{equation}
V^\pi(s) = \sum_{a} \pi(a|s) \left[ R(s,a) + \gamma \sum_{s'} P(s'|s,a) V^\pi(s') \right]
\end{equation}

The optimal value function $V^*(s)$ is defined as:

\begin{equation}
V^*(s) = \max_\pi V^\pi(s)
\end{equation}

and thus 

\begin{equation}
V^*(s) = \max_{a} \left[ R(s,a) + \gamma \sum_{s'} P(s'|s,a) V^*(s') \right]
\end{equation}

To estimate these value functions, various algorithms are employed. One example is Temporal Difference (TD) learning, where the estimate is updated incrementally:

\begin{equation}
V(S_t) \leftarrow V(S_t) + \alpha \left[ R_{t+1} + \gamma V(S_{t+1}) - V(S_t) \right]
\end{equation}

where $\alpha$ is the learning rate. Policy gradient methods can then optimize the policy directly by adjusting parameters $\theta$ of $\pi_\theta(a|s)$ to maximize expected returns.

\subsection{Behavioral Cloning}

Behavioral Cloning (BC) is a foundational approach within imitation learning, where an agent learns a policy by mimicking expert behavior through supervised learning. Rather than exploring the environment and receiving feedback via reward signals, as in traditional reinforcement learning (RL), BC directly maps observed states to expert actions using a dataset of demonstrations $\mathcal{D} = \{(s_i, a_i)\}_{i=1}^{N}$ collected from an expert policy $\pi_E$.

The learning objective in BC is to find a policy $\pi_\theta$ parameterized by $\theta$ that minimizes a supervised loss function over the dataset:

\begin{equation}
\mathcal{L}(\theta) = \frac{1}{N} \sum_{i=1}^{N} \ell\left(\pi_\theta(s_i), a_i\right)
\end{equation}

where $\ell$ is typically the cross-entropy loss for discrete actions or mean squared error for continuous control.

Although BC is simple and efficient to implement, especially in high-dimensional domains such as vision-based control \cite{bojarski2016end}, it suffers from the issue of covariate shift. Because the learned policy may encounter states that differ from those in the expert dataset, small prediction errors can compound over time, leading the agent to regions that are unseen or poorly represented in the state space \cite{ross2010efficient}. This phenomenon can severely degrade performance.

To address this, methods like Dataset Aggregation (DAgger) have been proposed, which iteratively augment the dataset with new trajectories generated by the learned policy and relabeled by the expert \cite{ross2011reduction}.

Despite its limitations, BC remains a strong baseline and is particularly useful in domains where safe exploration is critical or when expert demonstrations are abundant.

\subsection{Agentarena}
Agentarena is a 2D shooter game developed by Thomas Ackermann. The objective of the game player is to defeat the enemy in the agent arena by shooting at them. The arena is 900 by 1200 pixels large and can be seen in the following figure.

\begin{figure}[H]
    \centering
    \includegraphics[width=0.85\linewidth]{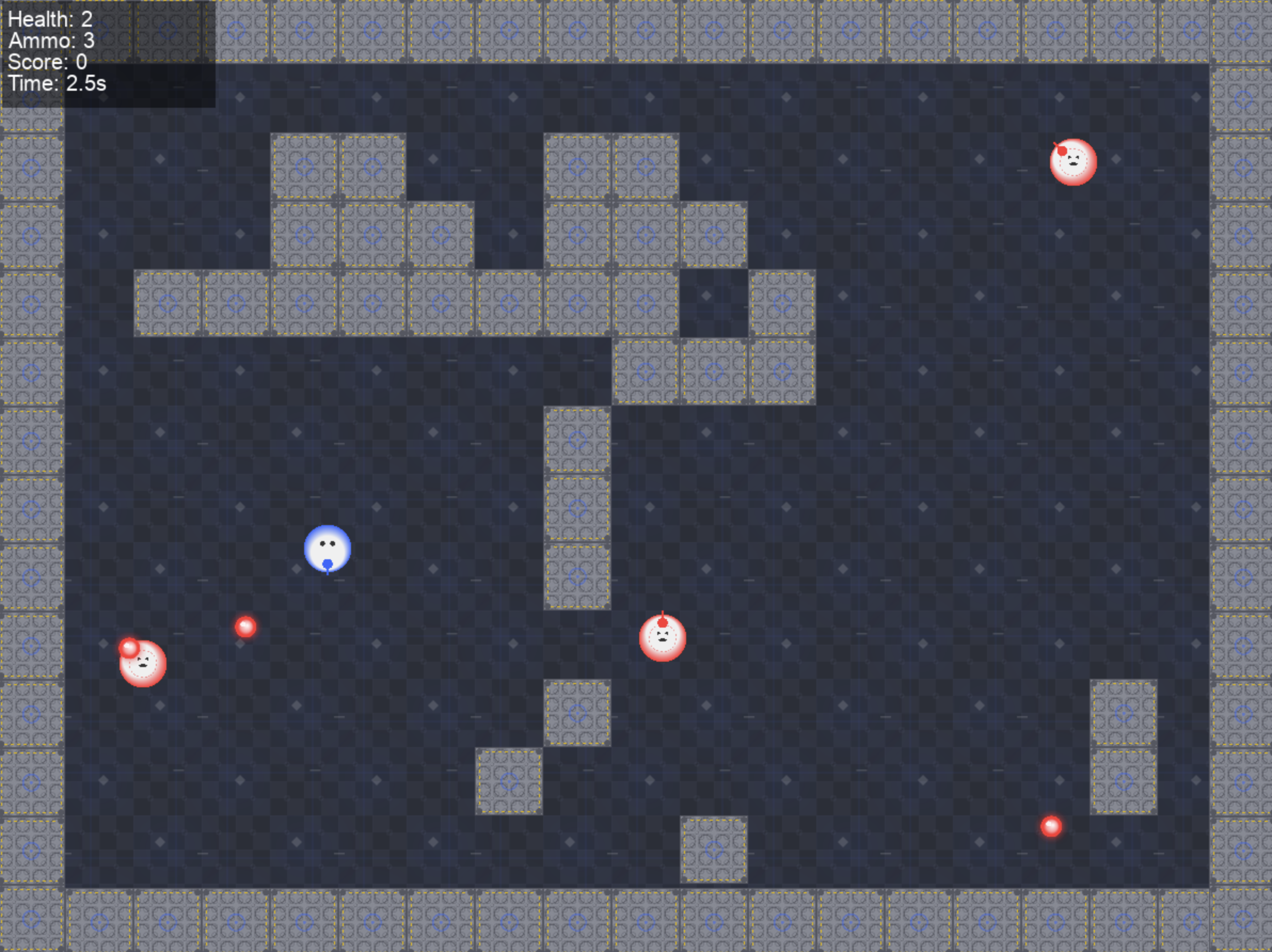}
    \caption{Playing field of the Agent arena game, the rectangular arena is obstructed by randomly placed walls that serve as cover for the player and enemies. The blue circle highlights the player entity, while the red circles highlight the enemy entities.}
    \label{fig:agentarena_arena}
\end{figure}
The game allows for the identification of state-space parameters such as the $x$ and $y$ coordinates of the player and enemy entities, whether the entities have fired a bullet or been hit. When an entity has been hit 3 times, it is removed from the game. When all enemies are eliminated this counts as a win. If the player dies, this counts as a loss.

\section{Related Work}

Recent research in reinforcement learning for shooter environments has increasingly turned toward hybrid approaches that integrate behavioral cloning (BC) with reinforcement learning (RL), especially to address instability and sparse reward signals observed in complex game environments. These methods are directly applicable to our 2D shooter game context, where rule-based strategies alone are insufficient and pure exploration leads to performance collapse. \\\\
Goecks et al.~\cite{goecks2019integrating} introduced the Cycle-of-Learning (CoL) framework, which combines BC with off-policy Q-learning to facilitate smooth transitions between imitation and exploration. Their multi-phase training pipeline begins with offline learning on expert data and progressively incorporates online RL. The authors demonstrate improved learning efficiency and stability in both dense and sparse reward environments, validating the value of hybrid loss designs for complex domains. 

Reddy et al.~\cite{reddy2019sqil} propose Soft Q Imitation Learning (SQIL), a method that injects expert demonstrations into the RL framework by assigning a fixed positive reward to expert actions and zero to all others. This simplifies integration by eliminating the need for separate BC objectives, yet still guides policy improvement through reinforcement. While not explicitly multi-headed, the principle of reward shaping via demonstrations shares conceptual overlap with alternating training phases.

Similarly, Spick et al. \cite{Spick.08.01.2024} adopted a BC-based approach to initialize the behavior of their FPS agent, aiming to rapidly and accurately mimic human gameplay, in contrast to strategies relying solely on RL. In their work, BC is subsequently augmented with RL to enable exploration and the acquisition of new skills over extended training periods. This hybrid, iterative framework introduces the challenge of appropriately balancing the influence of BC and RL. Fine-tuning this balance is critical to mitigate issues such as agent instability (e.g., becoming stuck) while still enabling the emergence of novel behaviors not captured by the initial BC model.

More recently, Torabi et al.~\cite{torabi2018behavioral} introduced a method for policy improvement through BC in environments where access to expert actions is limited. They show that even in such constrained scenarios, BC can produce a strong initial policy which, when refined through RL, exceeds the performance of either method in isolation.\\\\
These studies collectively affirm the advantages of combining imitation and reinforcement learning strategies. Hence, the research question is, how these strategies can be implemented for the 2D shooting game in a way that maximizes the win rate of the agent. Our work builds on this body of research by implementing a multi-head neural architecture that decouples imitation and Q-learning outputs, allowing gradient isolation and modular training schedules. In contrast to prior work, we apply these techniques to a custom-built 2D shooter environment with discrete action spaces, demonstrating that hybrid BC+RL strategies generalize beyond large-scale 3D simulations and into lightweight yet strategically rich game settings.

\section{Methods}

The development of an effective reinforcement learning agent for the 2D shooter game required a systematic approach combining multiple methodologies. This section details the implementation strategies used, from initial pure reinforcement learning attempts to the final training approach that combines offline learning with online reinforcement learning.

\subsection{Game Environment: AgentArena}

AgentArena serves as the testing environment for our reinforcement learning experiments. The game features a 2D top-down perspective where the player agent must eliminate enemy agents while avoiding being eliminated itself. The environment provides several key characteristics essential for RL training:

\textbf{State Space:} The observation space consists of player position and health, enemy positions and health, bullet positions and directions, and wall positions. This results in a high-dimensional state vector that captures all relevant game information.

\textbf{Action Space:} The agent can choose from $18$ discrete actions, combining $9$ movement directions (including staying still) with 2 shooting options (shoot or don't shoot), resulting in 18 total possible actions.

\textbf{Reward Structure:} The environment supports multiple reward functions of varying complexity, from basic hit/miss rewards to advanced tactical considerations including positioning, ammunition management, and survival strategies.

\subsection{Neural Network Architecture}

Our approach utilizes a multi-head neural network architecture designed to support both offline imitation learning and online reinforcement learning within a single model. The architecture consists of several key components:

\textbf{Feature Embedding Layers:} Separate embedding networks process different entity types (player, enemies, bullets, walls) with layer normalization and \mintinline{Python}{LeakyReLU} activations. This allows the model to learn specialized representations for each type of game entity.

\textbf{Attention Mechanisms:} Multi-head attention layers aggregate information from variable numbers of entities, using the player state as a query and other entities as keys and values. This enables the model to focus on relevant entities while handling variable numbers of enemies and bullets.

\textbf{Dual Output Heads:} The network features two separate output heads - a $Q$-learning head for reinforcement learning that outputs $Q$-values for action selection, and an imitation learning head for behavioral cloning that outputs action probabilities for supervised learning.

The complete network architecture processes input features through embedding layers, applies attention mechanisms for entity aggregation, and produces outputs through task-specific heads depending on the training mode.

\subsection{Training Methodology Evolution}

Our training approach evolved through several iterations based on empirical results and stability considerations.

\subsubsection{Initial Pure Reinforcement Learning}

The initial approach employed standard Deep $Q$-Network (DQN) training with experience replay and $\epsilon$-greedy exploration. However, this method encountered significant stability issues:

\textbf{Sparse Rewards:} The agent received rewards very infrequently, leading to slow learning and poor sample efficiency. Early reward functions only provided feedback for successful hits or eliminations, resulting in long periods without learning signals.

\textbf{Policy Instability:} Even when the agent achieved good performance, it would frequently revert to poor policies, indicating unstable learning dynamics. This manifested as inconsistent win rates and sudden drops in performance after periods of improvement.

\textbf{Scalability Issues:} The complexity of the multi-enemy environment made pure exploration inefficient, particularly when facing multiple opponents simultaneously.

\subsubsection{Reward Function Development}

To address the sparse reward problem, we developed progressively more sophisticated reward functions:

\textbf{Basic Rewards:} Simple hit/miss rewards with bonuses for enemy elimination and penalties for taking damage.

\textbf{Advanced Rewards:} Incorporated tactical considerations such as ammunition management, positioning relative to enemies, wall collision penalties, and bullet dodging rewards. The advanced reward function uses hyperbolic tangent normalization to bound reward values:

\begin{equation}
R_{advanced} = \tanh(R_{events} + R_{tactical} + R_{positional})
\end{equation}

where $R_{events}$ represents event-based rewards, $R_{tactical}$ captures tactical decision-making, and $R_{positional}$ encourages strategic positioning. The normalization function was chosen to limit the neural network output range.

\subsubsection{Introduction of Offline Learning}

Due to the instability of pure reinforcement learning, we introduced offline learning using behavioral cloning:

\textbf{Data Collection:} We collected demonstration data by recording gameplay from rule-based agents playing against various opponents. This provided a large dataset of state-action pairs representing competent gameplay. Initially we started out by using gameplay collected by human players. It turned out that the dataset was highly biased concerning the chosen actions.

\textbf{Behavioral Cloning:} The imitation learning head was trained using supervised learning on the demonstration data, learning to predict expert actions given game states. Cross Entropy was chosen as the loss function for imitation learning:

\begin{equation}
\mathcal{L}_{BC} = \mathbb{E}_{(s,a) \sim D} [\text{CrossEntropy}(\pi_\theta(s), a)]
\end{equation}

where $D$ represents the demonstration dataset and $\pi_\theta(s)$ represents the action probabilities from the imitation head.

\textbf{Action Balancing:} To address bias in the demonstration data, we implemented weighted sampling to balance the distribution of actions during training, preventing over-representation of common actions like movement without shooting.

\subsection{Hybrid Training Approach}

The final methodology combines offline and online learning in a dynamic schedule:

\subsubsection{Training Schedule Design}

The hybrid approach alternates between offline and online training episodes according to a decay schedule:

\begin{equation}
r_\text{offline}(t) = r_\text{initial} + \frac{t}{T} \cdot (r_{final} - r_\text{initial})
\end{equation}

where $r_\text{offline}(t)$ is the offline training ratio at episode $t$, $T$ is the total number of training episodes, $r_\text{initial}$ is the initial offline ratio (typically $0.8$), and $r_{final}$ is the final offline ratio (typically $0.2$).

\subsubsection{Multi-Head Training Protocol}

During offline episodes, the imitation learning head is trained using behavioral cloning on demonstration data. During online episodes, the $Q$-learning head is trained using DQN with experience replay. Both heads share the same feature extraction layers, allowing knowledge transfer between the two learning modes.

\textbf{Phase-Based Training:} Episodes are organized into phases of fixed length (typically $50$ episodes), with the offline ratio determining how many episodes in each phase use offline versus online training.

\textbf{Gradient Isolation:} The two training modes use separate optimizers and loss functions to prevent interference between imitation learning and reinforcement learning objectives.

\subsection{Experimental Design}

\subsubsection{Environment Configuration}

Experiments were conducted with standardized environment settings: single enemy configuration to reduce complexity, fixed arena size of $1200\times 900$ pixels, maximum episode length of $1000$ steps, and bullet speed and player movement speed optimized for balanced gameplay.

\subsubsection{Training Parameters}

Standard hyperparameters were established across experiments: learning rate of $0.001$ with step decay, discount factor $\gamma$ of $0.99$, initial exploration rate $\epsilon$ of $0.8$ decaying to $0.1$, batch size of $512$ for offline training, and experience replay buffer size of $20,000$ transitions.

\subsubsection{Evaluation Metrics}

Performance was measured using multiple metrics: win rate against rule-based opponents, average episode reward, episode length, accuracy (successful hits per shot fired), and training stability (variance in performance over time).

This comprehensive methodology addresses the key challenges of training reinforcement learning agents in complex game environments while maintaining training stability and achieving consistent performance improvements.

\section{Results}

To evaluate the effectiveness of our hybrid training approach, we conducted several experiments comparing different agent architectures and training configurations. Our evaluation framework tested trained agents against three distinct opponent types: a random agent that selects actions with equal probability, and two rule-based agents with varying complexity levels.

\subsection{Experimental Setup}

All models were initialized using a pretrained agent trained for 300 epochs through behavioral cloning on demonstration data collected from approximately $200$ episodes of rule-based gameplay. Here two rule-based agents competed against each other and the game observations and the respective actions of the agent was collected. The pretraining duration was selected based on validation performance monitoring, as shown in Figure \ref{fig:offline_validation_loss}, where we observed increasing validation error beyond $300$ epochs, indicating potential overfitting. Subsequently, each model underwent $1000$ episodes of online reinforcement learning training. This episode count was chosen based on computational constraints and empirical observation that win rates plateaued after extended training periods, as demonstrated in Figure \ref{fig:online_training_winrate}.

We evaluated three distinct model configurations to assess the impact of network architecture and exploration parameters:
\begin{itemize}
    \item Large neural network with initial exploration rate $\epsilon_0 = 0.4$
    \item Large neural network with initial exploration rate $\epsilon_0 = 0.8$ 
    \item Small neural network with initial exploration rate $\epsilon_0 = 0.8$
\end{itemize}

\subsection{Training Phase Analysis}

\begin{figure}[htbp]
    \centering
    \includegraphics[width=0.4\textwidth]{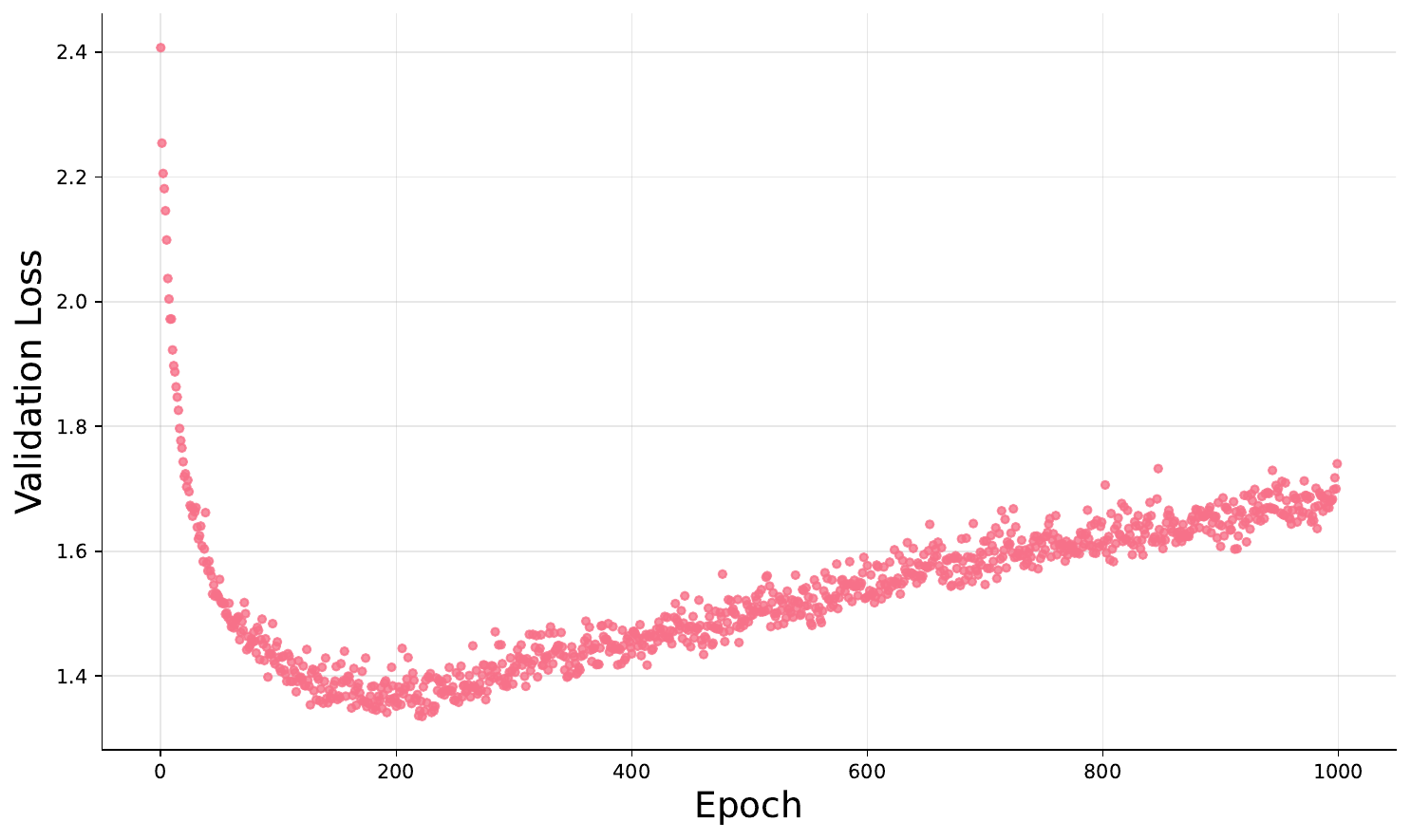}
    \caption{Validation loss during imitation learning pretraining. The loss decreases steadily until approximately 300 epochs, after which it begins to rise, indicating overfitting. This guided our selection of the pretraining duration.}
    \label{fig:offline_validation_loss}
\end{figure}

Figure \ref{fig:offline_validation_loss} illustrates the validation loss trajectory during the behavioral cloning phase. The model demonstrates clear learning progress with decreasing validation loss until approximately $300$ epochs, after which the loss begins to increase, signaling the onset of overfitting. This empirical evidence justified our decision to terminate pretraining at $300$ epochs to preserve generalization capability.

\begin{figure}[htbp]
    \centering
    \includegraphics[width=0.4\textwidth]{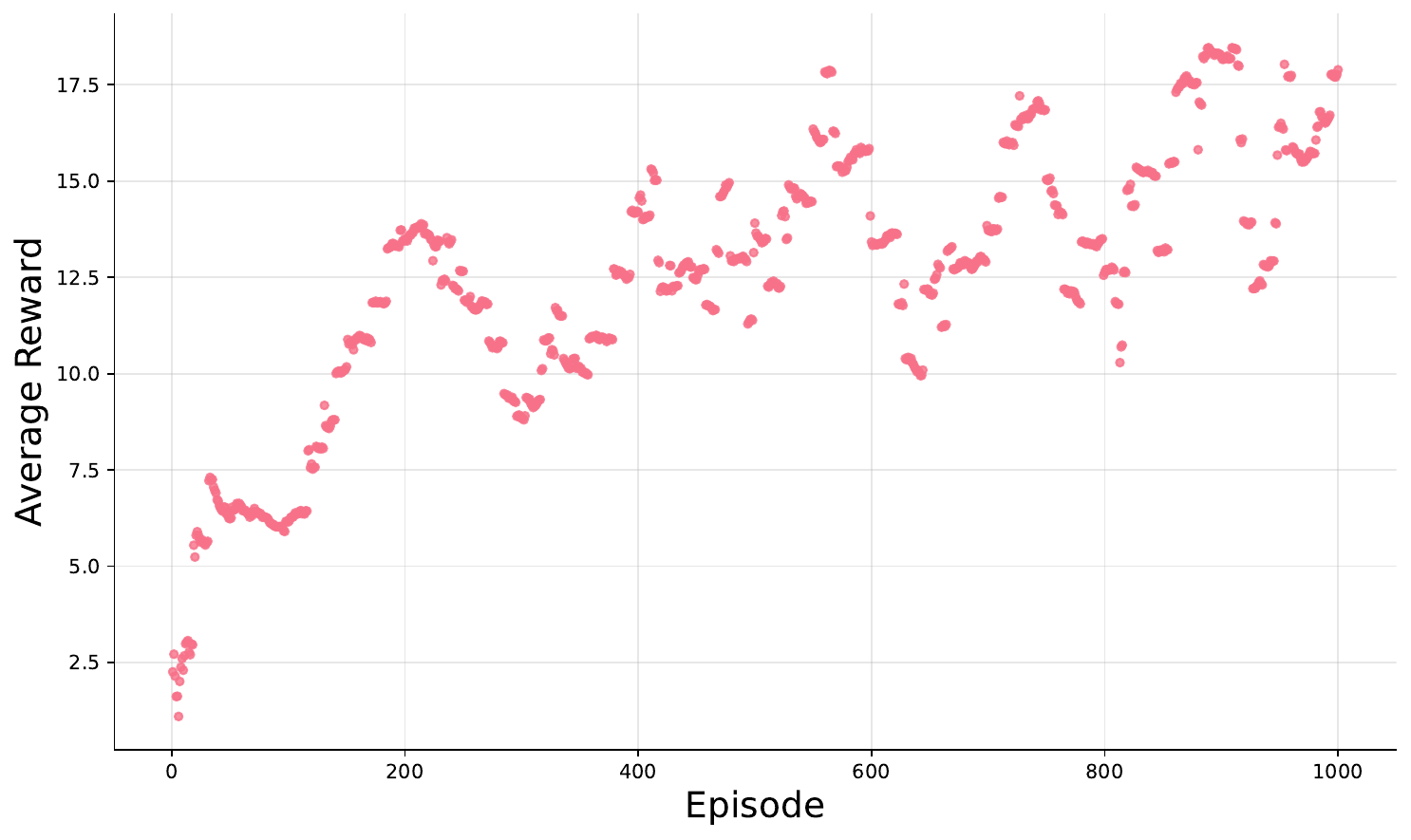}
    \caption{Average reward progression during online reinforcement learning training, showing steady improvement and eventual convergence.}
    \label{fig:online_average_reward}
\end{figure}

\begin{figure}[htbp]
    \centering
    \includegraphics[width=0.4\textwidth]{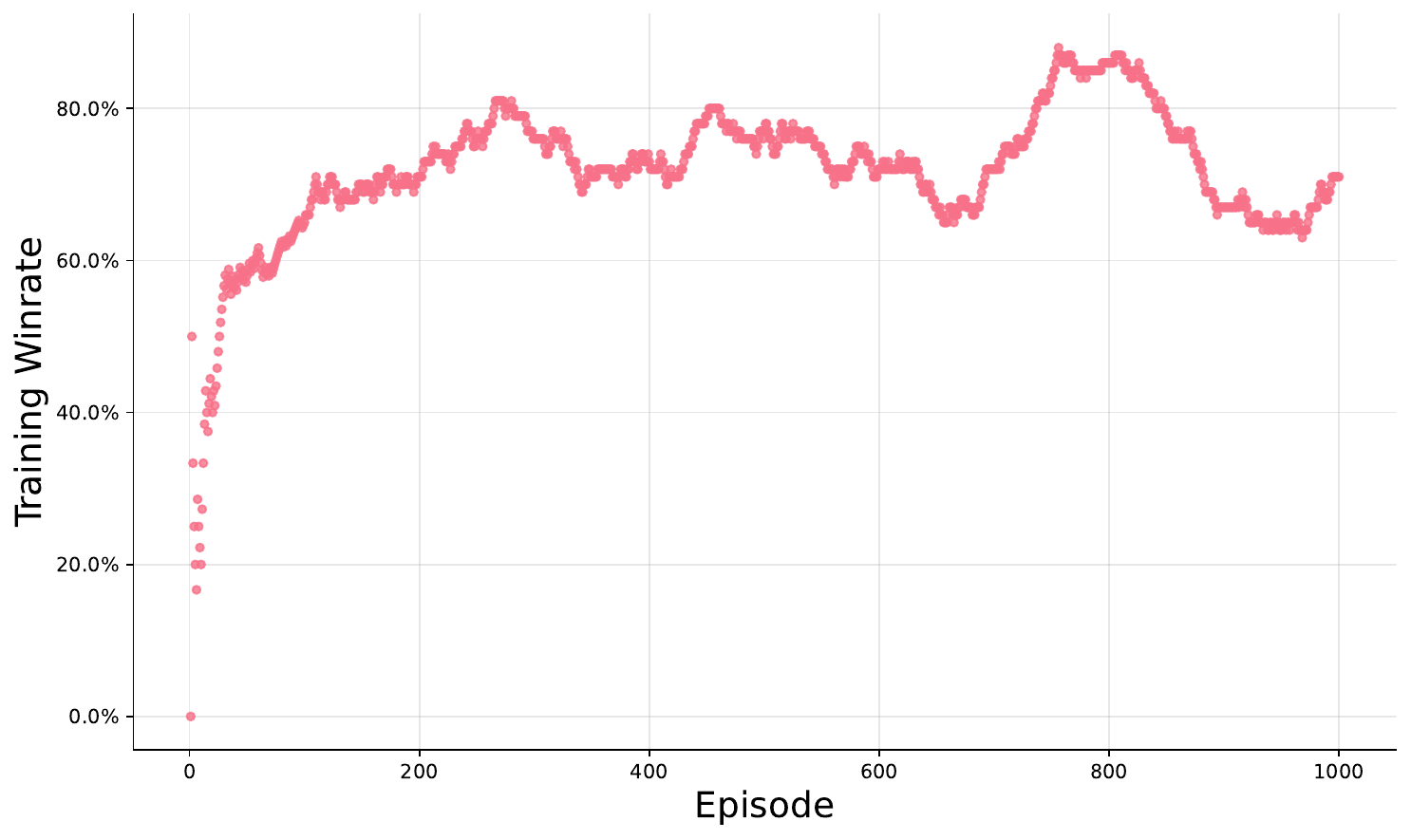}
    \caption{Win rate evolution during reinforcement learning training. The initially low win rate rapidly improves and stabilizes at a high level, demonstrating the effectiveness of pretraining in providing a stable foundation for RL optimization.}
    \label{fig:online_training_winrate}
\end{figure}

The online training phase, depicted in Figures \ref{fig:online_average_reward} and \ref{fig:online_training_winrate}, demonstrates the effectiveness of our hybrid approach. The average reward shows steady improvement throughout training (Figure \ref{fig:online_average_reward}), while the win rate exhibits rapid initial improvement followed by stable high performance (Figure \ref{fig:online_training_winrate}). Notably, the win rate begins at a reasonable baseline due to pretraining, avoiding the typically poor initial performance associated with pure reinforcement learning approaches.

\subsection{Performance Against Different Opponents}

\begin{table}[htbp]
\centering
\begin{tabular}{lcc} 
\toprule
\textbf{Enemy Type} & \textbf{Win Rate (\%)} & \textbf{Avg Episode Length (steps)} \\ 
\midrule
Rule Based & 76.0 & 117.8 \\
Rule Based 2 & 86.0 & 180.5 \\
Random & 72.0 & 117.6 \\ 
\bottomrule
\end{tabular}
\caption{Performance of large neural network architecture with starting exploration rate $\epsilon_0 = 0.4$.}
\label{tab:results_large_04}
\end{table}

\begin{table}[htbp]
\centering
\begin{tabular}{lcc} 
\toprule
\textbf{Enemy Type} & \textbf{Win Rate (\%)} & \textbf{Avg Episode Length (steps)} \\ 
\midrule
Rule Based & 94.0 & 73.3 \\
Rule Based 2 & 76.0 & 264.2 \\
Random & 66.0 & 129.8 \\ 
\bottomrule
\end{tabular}
\caption{Performance of large neural network architecture with starting exploration rate $\epsilon_0 = 0.8$.}
\label{tab:results_large_08}
\end{table}

\begin{table}[htbp]
\centering
\begin{tabular}{lcc} 
\toprule
\textbf{Enemy Type} & \textbf{Win Rate (\%)} & \textbf{Avg Episode Length (steps)} \\ 
\midrule
Rule Based & 96.0 & 105.3 \\
Rule Based 2 & 92.0 & 97.3 \\
Random & 78.0 & 207.0 \\ 
\bottomrule
\end{tabular}
\caption{Performance of small neural network architecture with starting exploration rate $\epsilon_0 = 0.8$.}
\label{tab:results_small_08}
\end{table}

\subsection{Performance Analysis}

\subsubsection{Superior Performance Against Training Source}

Most notably, all model configurations achieved win rates of 76-96\% against the Rule Based agent, which served as the source of demonstration data for pretraining. This indicates that the hybrid approach successfully enabled agents to surpass their initial teachers, validating the effectiveness of combining offline learning with online optimization.

\subsubsection{Exploration Rate Impact}

Comparing the large network configurations reveals the critical importance of exploration strategy. Higher initial exploration ($\epsilon_0 = 0.8$) yielded dramatically improved performance against the Rule Based agent (94\% vs 76\% win rate) and significantly shorter episodes (73.3 vs 117.8 steps). This suggests that increased exploration during early training phases allows agents to discover more efficient strategies beyond those demonstrated in the training data, leading to both higher success rates and more decisive victories.

\subsubsection{Network Architecture Effects}

Contrary to expectations, the small neural network architecture outperformed the large architecture in several metrics. The small network achieved the highest win rates against Rule Based (96\%) and Rule Based 2 (92\%) opponents, while maintaining competitive performance against Random opponents (78\%). This finding suggests that architectural design and training methodology are more critical than raw network capacity for this particular task domain. Another reason could also be that we ran fast into overfitting during the BC phase. Most likely there was not enough data or the data was too similar as we trained against the same agent.

\subsubsection{Opponent-Specific Performance Patterns}

The results reveal distinct performance patterns across opponent types that provide insights into the nature of learned strategies:

\textbf{Rule-Based Opponents:} Agents achieved consistently high performance (76-96\% win rates) against both rule-based opponents, with the small network showing particularly strong results. The varying episode lengths (73.3-264.2 steps) suggest that agents adapted their strategies to exploit specific opponent patterns efficiently.

\textbf{Random Opponents:} Performance against random agents remained lower across all configurations (66-78\% win rates), which can be attributed to the inherently unpredictable nature of random behavior. Random agents exhibit erratic patterns that cannot be easily learned or exploited, making them challenging opponents despite their lack of strategic planning. The longer episode lengths against random opponents (129.8-207.0 steps) indicate longer duels due to the unpredictable patterns of the opponent.

\subsubsection{Episode Length Analysis}

The variation in episode lengths across configurations provides additional insights into tactical efficiency:
\begin{itemize}
    \item \textbf{Short episodes} (73.3-117.8 steps) against Rule Based agents suggest efficient exploitation of predictable patterns
    \item \textbf{Extended episodes} (180.5-264.2 steps) against Rule Based 2 agents indicate more complex strategic interactions
    \item \textbf{Variable lengths} against Random opponents reflect the unpredictable nature of random behavior
\end{itemize}

\subsection{Training Stability and Convergence}

The hybrid training approach demonstrated superior stability compared to pure reinforcement learning methods encountered in preliminary experiments. The pretraining phase provided a stable foundation that prevented the catastrophic forgetting and policy collapse commonly observed in pure RL approaches for this domain.

As evidenced in Figure \ref{fig:online_training_winrate}, the win rate begins at a reasonable baseline due to behavioral cloning initialization, then rapidly improves and stabilizes at high performance levels. This contrasts sharply with typical pure RL training curves that often exhibit extended periods of poor performance before achieving competency.

The combination of behavioral cloning initialization followed by reinforcement learning optimization proved effective in achieving both stability and performance improvement. Agents successfully learned to exploit opponent patterns while developing tactical behaviors beyond those present in the demonstration data, as evidenced by their ability to surpass the performance of their rule-based teachers across all tested configurations.

\section{Conclusion}
This study presents a robust pretraining framework that effectively combines behavioral cloning (BC) with deep reinforcement learning (RL) to train a competent agent for a 2D shooter game environment. The proposed approach addresses the limitations commonly associated with pure RL—such as training instability, sparse rewards, and poor sample efficiency—by initializing the agent with a policy learned from expert demonstrations and subsequently refining it through online reinforcement learning.

Empirical evaluations demonstrate that the hybrid agent significantly outperforms agents trained solely through reinforcement learning, achieving win rates exceeding 90 \% against rule-based opponents. Furthermore, the integration of a multi-head neural network architecture with shared attention-based feature extraction facilitates seamless knowledge transfer between the imitation and reinforcement learning modalities. This contributes to improved training stability and convergence behavior.

Notably, the hybrid training schedule—gradually transitioning from offline to online learning—proved critical in balancing exploration with policy retention. Experiments also highlighted the importance of early-phase exploration parameters in enabling agents to discover more efficient strategies beyond those seen in demonstration data.

Overall, the findings affirm the effectiveness of leveraging demonstration-driven policy initialization as a foundation for RL optimization in complex multi-agent environments. Future work may focus on incorporating curriculum learning, dynamic reward shaping, and multi-agent coordination to further enhance agent adaptability and performance in increasingly sophisticated scenarios.

\ifCLASSOPTIONcaptionsoff
  \newpage
\fi

\bibliographystyle{IEEEtran}

\begin{thebibliography}{10}
\providecommand{\url}[1]{#1}
\csname url@samestyle\endcsname
\providecommand{\newblock}{\relax}
\providecommand{\bibinfo}[2]{#2}
\providecommand{\BIBentrySTDinterwordspacing}{\spaceskip=0pt\relax}
\providecommand{\BIBentryALTinterwordstretchfactor}{4}
\providecommand{\BIBentryALTinterwordspacing}{\spaceskip=\fontdimen2\font plus
\BIBentryALTinterwordstretchfactor\fontdimen3\font minus \fontdimen4\font\relax}
\providecommand{\BIBforeignlanguage}[2]{{%
\expandafter\ifx\csname l@#1\endcsname\relax
\typeout{** WARNING: IEEEtran.bst: No hyphenation pattern has been}%
\typeout{** loaded for the language `#1'. Using the pattern for}%
\typeout{** the default language instead.}%
\else
\language=\csname l@#1\endcsname
\fi
#2}}
\providecommand{\BIBdecl}{\relax}
\BIBdecl

\bibitem{sutton2018reinforcement}
R.~S. Sutton and A.~G. Barto, \emph{Reinforcement Learning: An Introduction}.\hskip 1em plus 0.5em minus 0.4em\relax MIT Press, 2018.

\bibitem{li2017deep}
Y.~Li, ``Deep reinforcement learning: An overview,'' \emph{arXiv preprint arXiv:1701.07274}, 2017.

\bibitem{kalashnikov2018qt}
D.~Kalashnikov \emph{et~al.}, ``Qt-opt: Scalable deep reinforcement learning for vision-based robotic manipulation,'' \emph{arXiv preprint arXiv:1806.10293}, 2018.

\bibitem{komorowski2018artificial}
M.~Komorowski \emph{et~al.}, ``The artificial intelligence clinician learns optimal treatment strategies for sepsis in intensive care,'' \emph{Nature Medicine}, vol.~24, pp. 1716--1720, 2018.

\bibitem{bellemare2013arcade}
M.~G. Bellemare, Y.~Naddaf, J.~Veness, and M.~Bowling, ``The arcade learning environment: An evaluation platform for general agents,'' in \emph{Journal of Artificial Intelligence Research}, 2013.

\bibitem{kempka2016vizdoom}
M.~Kempka \emph{et~al.}, ``Vizdoom: A doom-based ai research platform for visual reinforcement learning,'' in \emph{IEEE Conference on Computational Intelligence and Games (CIG)}, 2016.

\bibitem{silver2016mastering}
D.~Silver \emph{et~al.}, ``Mastering the game of go with deep neural networks and tree search,'' \emph{Nature}, vol. 529, no. 7587, pp. 484--489, 2016.

\bibitem{silver2018general}
------, ``A general reinforcement learning algorithm that masters chess, shogi, and go through self-play,'' \emph{Science}, vol. 362, no. 6419, pp. 1140--1144, 2018.

\bibitem{bojarski2016end}
M.~Bojarski \emph{et~al.}, ``End to end learning for self-driving cars,'' 2016.

\bibitem{ross2010efficient}
S.~Ross, G.~Gordon, and D.~Bagnell, ``A reduction of imitation learning and structured prediction to no-regret online learning,'' in \emph{Proceedings of the thirteenth international conference on artificial intelligence and statistics}, 2010, pp. 627--635.

\bibitem{ross2011reduction}
S.~Ross and J.~A. Bagnell, ``A reduction of imitation learning and structured prediction to no-regret online learning,'' in \emph{Proceedings of the Fourteenth International Conference on Artificial Intelligence and Statistics}, 2011, pp. 627--635.

\bibitem{goecks2019integrating}
V.~Goecks, A.~Lee, J.~Hays, J.~Valasek, and P.~Stone, ``Integrating behavior cloning and reinforcement learning for improved performance in dense and sparse reward environments,'' in \emph{Proceedings of the International Conference on Autonomous Agents and MultiAgent Systems}, 2019.

\bibitem{reddy2019sqil}
S.~Reddy, A.~Dragan, and S.~Levine, ``Sqil: Imitation learning via regularized behavioral cloning,'' \emph{arXiv preprint arXiv:1905.11108}, 2019.

\bibitem{Spick.08.01.2024}
\BIBentryALTinterwordspacing
R.~Spick, T.~Bradley, A.~Raina, P.~V. Amadori, and G.~Moss, ``Behavioural cloning in vizdoom.'' [Online]. Available: \url{http://arxiv.org/pdf/2401.03993}
\BIBentrySTDinterwordspacing

\bibitem{torabi2018behavioral}
F.~Torabi, G.~Warnell, and P.~Stone, ``Behavioral cloning from observation,'' \emph{arXiv preprint arXiv:1805.01954}, 2018.

\end{thebibliography}

\end{document}